%% file: paper.tex
\documentclass{article}

\usepackage{PRIMEarxiv}

\usepackage[utf8]{inputenc} 
\usepackage[T1]{fontenc}    
\usepackage{hyperref}       
\usepackage{url}            
\usepackage{booktabs}       
\usepackage{amsfonts}       
\usepackage{nicefrac}       
\usepackage{microtype}      
\usepackage{lipsum}
\usepackage{fancyhdr}       
\usepackage{graphicx}       
\graphicspath{{media/}}     

\usepackage{tabularx}
\usepackage{adjustbox}
\usepackage{pdflscape}
\usepackage{geometry}
\usepackage{longtable}

\usepackage{amsmath}

\title{Advancing Crime Linkage Analysis with Machine Learning: A Comprehensive Review and Framework for Data-Driven Approaches
}

\author{
  Vinicius Lima\\
  Computer and Information Technology \\
  Purdue Univeristy \\
  West Lafayette\\
  \texttt{vlima@email} \\
   \And
  Umit Karabiyik\\
  Computer and Information Technology \\
  Purdue Univeristy \\
  West Lafayette\\
  \texttt{ukarabik@email} \\
}

\begin{document}
\maketitle

\begin{abstract}
Crime linkage is the process of analyzing criminal behavior data to determine whether a pair or group of crime cases are connected or belong to a series of offenses. This domain has been extensively studied by researchers in sociology, psychology, and statistics. More recently, it has drawn interest from computer scientists, especially with advances in artificial intelligence. Despite this, the literature indicates that work in this latter discipline is still in its early stages. This study aims to understand the challenges faced by machine learning approaches in crime linkage and to support foundational knowledge for future data-driven methods. To achieve this goal, we conducted a comprehensive survey of the main literature on the topic and developed a general framework for crime linkage processes, thoroughly describing each step. Our goal was to unify insights from diverse fields into a shared terminology to enhance the research landscape for those intrigued by this subject.
\end{abstract}


\section{Introduction}

Crime Linkage (CL) \cite{green1976cluster} is a multidisciplinary field that has garnered significant attention from sociologists, psychologists, statisticians, and computer scientists \cite{porter_statistical_2016,woodhams_psychology_2007,davies_practice_2019,fox_what_2018,zhu_spatiotemporal-textual_2022,li_supervised_2022}. In essence, they aim to connect pairs of crimes by creating crime link associations \cite{woodhams_psychology_2007}. The CL analysis is typically based on Modus Operandi (MO) or criminal behavior, which can be inferred from evidence found at crime scenes or other source of police data \cite{douglas_violent_1992}. These MOs/behaviors are systematically compared to assess the similarities between different crimes \cite{melnyk_another_2011}. High similarity scores often indicate potential connections, providing valuable insight into shared offender characteristics or underlying patterns within these criminal acts.

The literature on crime linkage encompasses a wide array of methods and evaluation metrics, alongside various data sets used to establish connections between criminal incidents. Numerous studies have showcased CL outcomes, particularly in cases involving property crimes and sexual offenses \cite{fox_what_2018}. The CL is in essence a classification problem, where the goal is to either assess if a pair of crimes is linked or not, or to associate a case to a series of offenses, based on pattern characteristics.

There are many methodologies for modeling and evaluating crime linkage. Data-driven approaches in CL have been explored using statistical methods, fuzzy logic, and machine learning techniques. Although recent years have seen a surge in artificial intelligence (AI) research, studies applying machine learning to crime linkage are still in the early stages. We sought to understand the reasons behind this. To gain insight, we surveyed key data-driven work, including both statistical and machine learning approaches, and analyzed the challenges - whether implicit or explicit - associated with handling large amounts of data.

Previous literature surveys have not included machine learning approaches \cite{woodhams_psychology_2007, davies_practice_2019, fox_what_2018}, thus, our aim is to provide an extension of research assessment on the theme. When analyzing the studies collected, we observed a general framework in the CL process, which will be also discussed in this paper. 

In summary, the contributions of this study is two fold:
\begin{itemize}
    \item (1) A comprehensive survey of the main literature on data-drive crime linkage and outline of a general framework of the linkage process to support further research in the topic.
    \item (2) Analyze the main challenges faced by crime linkage, especially when dealing with machine learning and bigger datasets, and thus support future work.
\end{itemize}

The structure of this paper unfolds as follows. First, we will discuss the main literature on crime linkage, outlining the concepts and definitions involved. Then, in Section 3, we will describe the general framework usually used to work on CL, as a discussion of each step and the techniques involved. Section 4 is dedicated to elaborate on the challenges faced by the topic, as potential solutions when applicable. Finally, we will wrap up with the main conclusions of this work. The Appendix (Section \ref{appendix})  will show all the papers surveyed and details on each study.

\section{Related Work}

Crime Linkage, a.k.a. Case Linkage or Behavioral Crime Linking, has been a subject of extensive study, particularly within domains such as Criminology and Psychology. Its origins trace back to the 1970s \cite{green1976cluster}, with a notable surge in practical research observed in the 2000s \cite{davies_practice_2019}. Although related to criminal profiling \cite{fox_what_2018}, crime linkage specifically refers to ``the process of linking two or more crimes together on the basis of the crime scene behavior exhibited by an offender'' \cite{woodhams_case_2007}.  In other words, this discipline is interested in understanding pattern characteristics in criminal behavior that might suffice connections between which seem to be unrelated crimes, often in the context of serial offenders.

Crime Linkage is underpinned by two assumptions in criminal behavior: consistency and distinctiveness. The latter posits that serial criminals exhibit patterned actions akin to a signature behavior \cite{canter_psychology_1995}, while the former suggests that this signature is unique enough to differentiate their criminal activity from others \cite{woodhams_case_2007}. However, in terms of data-driven CL, Bennell et al.  \cite{bennell_computerized_2012} introduced three other assumptions that challenge research in this area: the reliability of coded data within systems, the accuracy of the data, and analysts' ability to accurately link crimes using this data. Basically, the reliability and suitability of the data to draw conclusions on the link between crime remain the subject of debate \cite{martineau_investigating_2008}.

The primary focus of this topic lies in identifying pairs or series of crimes committed by the same perpetrator. This field is of significant importance, and CL has already been used as evidence in court proceedings to implicate suspects \cite{bosco_admissibility_2010, labuschagne_use_2006, meyer_criminal_2007}. In particular, Keppel documented the first case in Canada where crime linkage was used as evidence, which a decision later corroborated the result through DNA analysis~\cite{keppel_signature_2000}. However, the authors have debated whether and how pattern behavior can be used as evidence in court. Labuschagne~\cite{labuschagne2014use} points out that a linkage decision should not be based only on computerized approaches, while Canter et al.~\cite{canter2004organized} do not recommend exclusively experience-based decisions.

In 1993, the U.S. Supreme Court established guidelines for the incorporation of scientific evidence into legal proceedings, now enshrined in Rule 702, commonly known as the Daubert standard~\cite{pakkanen2014crime}. This landmark decision provided a framework to bolster the credibility of forensic science evidence. The standards outlined for admissible scientific evidence include factors such as reliability, peer review and publication, error rates, general acceptance within the scientific community, the existence of known standards, and applicability to the specific case at hand. In the context of Data-Driven Crime Linkage and its conformity with the Daubert standard, Pakkanen et al.~\cite{pakkanen2014crime} proposed an evaluation framework centered on three key aspects:

\begin{itemize}
    \item \textit{Consistent Behavior:} This pertains to assessing the probability of consistency in linking one crime to another and the associated margin of error. It also involves determining to what extent the findings from specific samples can be generalized to broader populations. A comprehensive examination of the core attributes linking ostensibly distinct crimes becomes imperative.
    \item \textit{Reliable Database:} This aspect revolves around the meticulous coding of variables in databases. Given the subjective nature of variables, careful selection methods are essential yet challenging. In addition, determining which variables are crucial in establishing crime linkages warrants further investigation. Furthermore, databases should encompass not only solved cases but also unsolved ones. Thus, preliminary studies are needed to discern when a crime constitutes a part of a serial set to ensure ecological validity.
    \item \textit{Frequency:} This involves quantifying the similarity between linked crimes. The mere presentation of similarities and differences in variables between crime pairs is insufficient; it is equally vital to ascertain the likelihood of attributing them to the same offender, particularly in cases involving serial and one-off occurrences.
\end{itemize} 

The utilization of databases to facilitate crime linkage traces back to the inception of the FBI's Violent Criminal Apprehension Program (ViCAP) in 1985 \cite{bennell_computerized_2012}. ViCAP was specifically designed to serve as a centralized intelligence hub aimed at identifying potential connections among seemingly disparate crimes, avoiding the problem known as linkage blindness \cite{egger1984working}. The program was developed to mitigate issues stemming from communication gaps between police stations and to facilitate the identification of links that span across different jurisdictions, especially those related to homicide, sexual offenses,and missing people \cite{egger_working_1984}. However, ViCAP relies primarily on the manual entry of relevant crime scene and MO data, requiring training and expertise to potentially provide links effectively. Another significant database employed for CL purposes is the Violent Crime Linkage Analysis System (ViCLAS), which has been used in many countries such as Canada, New Zealand, Germany, Belgium, Great Britain, and some American states \cite{collins_advances_1998}.

However, the coding process for entering information into these databases has been scrutinized. To demonstrate this, Martineau and Corey~\cite{martineau_investigating_2008} conducted a study testing the level of agreement among Canadian police officers in coding crime linkage scenarios, yielding a 38\% agreement for homicide cases and 25\% for sexual assault cases. This highlights the subjectivity of human evaluation on criminal matters, which can greatly impact the assessment of when a crime can be considered linked or not, potentially leading to false accusations. Moreover, the process of decision-making for crime linkage itself has sparked discussion due to the absence of standardized training or evaluation methods for this intricate and subjective decision-making process. For instance, Bennell et al.~\cite{bennell_linkage_2010} demonstrated that students outperformed police personnel in associating crimes with serial offenders, and a logistic regression model outperformed humans. 

The field of CL encompasses various methodologies, and some have divided them into Comparative Case Analysis (CCA) and Case Linkage Analysis (CLA)  \cite{davies_practice_2019}. CCA involves comparing features of a crime against a database to identify similar cases. A survey conducted by Burrell and Bull (2011) with crime analysts from the UK police highlighted the challenges in identifying series of crimes by analyzing elements of crime scene behavior, including time, location, \textit{modus operandi}, crime evidence, and characteristics of offenders and victims \cite{burrell_preliminary_2011}. However, gathering forensic data, particularly for CCA purposes, was noted as particularly challenging. Within CCA, two distinct sub-methods exist: reactive and proactive \cite{woodhams_case_2007}. Reactive Linkage involves comparing a specific index crime with many or all crimes in a database, often providing a similarity score to quantify the resemblance between two criminal activities or the likelihood of them being perpetrated by the same offender. Proactive Linkage, on the other hand, clusters crimes together to identify potential series of crimes.

CLA focuses on determining whether pairs of crimes or a small subset are linked \cite{rainbow_practitioners_2014}. Typically, experts offer opinions based on their own methodologies to establish links between cases, relying on specific signature behaviors repeatedly exhibited by the criminal \cite{schlesinger_serial_2000, keppel_serial_2008}. While experts may provide detailed information on MO characteristics, these are often specific to particular cases and cannot be generalized \cite{davies_practice_2019}. Notably, famous serial killers have been studied and analyzed based on their signature behaviors, enabling the connection of their crimes \cite{keppel_jack_2005,yokota_practice_2017,labuschagne_use_2006}. 

\subsection{\textit{Modus Operandi} in the Crime Linkage Context}

As highlighted, understanding signature criminal behavior is pivotal to crime linkage. Commonly known as \textit{modus operandi}, it refers to the patterned behavior exhibited by a perpetrator during criminal activities, aimed at safeguarding their identity, selecting a victim, ensuring the success of the crime, facilitating escape, and evading detection \cite{cornish_understanding_2017, douglas_crime_2012, gee_sex_2014, hazelwood_relevance_2016, leclerc_examining_2009}. In the context of crime linkage, the aim is to gather sufficient evidence to discern these behavioral patterns and encapsulate them as a signature behavior. Assuming they are consistent and ``individualizable'' (distinctiveness), these patterns can be utilized to identify serial offenders or establish connections between related crimes (for example, using MOs to cluster a network of offenders). Data-driven studies typically deal with a predefined set of MO attributes, usually coded by experts or directly inputted into databases. However, this approach somewhat contradicts the assumptions posited by Woodhams et al. (2007)  \cite{woodhams_psychology_2007} since a finite and discrete number of MO characteristics may limit the diverse possibilities of distinctiveness within offender behavior. It is expected that the signature behavior lies within the nuances of the crime scene rather than in a generalized combination of attributes. This will be further discussed in the challenges section.

The discussion of whether MO is sufficient to translate links in crime has been topic of many research discussions. In fact, certain studies have indicated that spatial analysis of crime (i.e. proximity of crimes) is a more effective feature for linking crimes compared to traditional MO \cite{bennell_between_2005}. Indeed, due to the dynamic nature of MO, extracting its key elements is a hard task that necessitates extensive experience \cite{douglas_violent_1992}. This difficulty in delineating MO may contribute to the scarcity of data-driven studies in crime linkage, particularly in homicides \cite{fox_what_2018}, as it will be discussed later. It can be inferred that burglaries and robberies can often exhibit a well-defined set of MO attributes, while the same cannot be said for other type of crime.

\subsection{Data-driven studies}

This section is dedicated to briefly outline the main applied studies on Crime Linkage, with a particular emphasis on data-driven methods, i.e. falling under the CCA approach. Four literature review papers were identified concerning crime linkage in practice. Woodhams et al. \cite{bennell2012computerized} provided an early review, discussing fundamental concepts and the primary challenges faced by CL at that time. Bennell et al. (2012) delved into the underlying assumptions of computerized crime linkage systems. In a subsequent work, authors in \cite{bennell2014linking} analyzed evaluation measures of crime linkage, particularly focusing on the AUC (Area Under the Curve), and identified crime type, behavior domain, and distance as key factors affecting performance. While predominantly focused on criminal profiling, Fox et al.~\cite{fox_what_2018} dedicated a section to discussing the 17 papers compiled by Bennell et al. (2014), with a focus on the effect size of the studied samples. Davies et al.~\cite{davies2019practice} conducted the most recent literature review on the topic, describing the challenges to obtain crime linkage in practice. Their focus extended beyond the methodology for crime linkage outcomes to encompass studies exploring the overall practice and usage of crime linkage, yielding 30 relevant papers. Simirlaly, this study concentrates specifically on practical methods for deriving crime linkage, particularly focusing on studies utilizing crime datasets. However, unlike previous works, this study will emphasize statistical/machine learning methods and the nuances of applying these techniques.

In this context, much of the research has focused on property crimes such as burglary \cite{bennell2002linking, bennell_between_2005, markson2010linking, melnyk2011another, tonkin2012linking, tonkin2012comparison, zhu_spatial-temporal-textual_2021, zhu_spatiotemporal-textual_2022, solomon_crime_2020, ku2014decision, reich2015partially, albertetti2013crilim, qazi2019interactive, porter_statistical_2016, yokota2002computer}, robbery \cite{woodhams2007empirical, burrell2012linking, li_approach_2019, zhu_spatial-temporal-textual_2021, li_novel_2020, li_supervised_2022-1, chi_decision_2017}, car theft \cite{tonkin2008link, davies2012course, tonkin2012comparison}, and arson \cite{ellingwood2013linking}. In terms of crimes against individuals, there is a notable focus on sexual crimes (not necessarily involving murder) \cite{bennell2009addressing, bennell2010impact, woodhams2012test, winter2013comparing, oziel2015variability, yokota2017crime, slater2015testing, tonkin2017using}, with three studies delving into homicide \cite{melnyk2011another, salo2013using, santtila2008behavioural}. Additionally, there are studies where researchers have examined crime linkage across various crime types, including a variety of crime categories \cite{tonkin2011linking}, burglaries, robberies, and car thefts \cite{tonkin2012behavioural, tonkin_linking_2019}, and burglaries, robberies, and assaults in general \cite{ku2014decision}. Refer to the Appendix table (Section \ref{appendix}) to check other crime types related to topic in question in this paper.


Within purely statistical approaches, researchers have explored various methodologies to ascertain the probability of linking one crime to another, employing techniques such as logistic regression (LR), Naive Bayes, and decision trees (DT). Naive Bayes is based on Bayes' Theorem, which calculates the probability of a particular event occurring given the occurrence of another event. Despite its simplicity, the "naive" assumption assumes that features (variables) are independent of one another, which rarely holds in real-world scenarios but still often yields good results. For example, Porter (2016) took this a step further by utilizing the Bayes factor to inform the decision-making process regarding which model to employ \cite{porter_statistical_2016}. The main advantage for using Bayes models is that the results are more easily interpret as compared to other more advanced machine learning techniques.

The main methodological preference among many researchers for CL is logistic regression~\cite{cox1958regression}. This approach is suitable for CL, as the task typically involves binary classification—deciding whether crimes are linked or not. LR estimates the probability of crimes being linked based on a linear combination of input features, using a logistic function (sigmoid, for instance) to constrain the output between 0 and 1, which can be interpreted as a probability. During training, the model learns the optimal weights for the features and identifies a decision threshold, which can later be evaluated for performance. One limitation of this method is its assumption of a linear relationship between the input features (crime features) and the output (linkage decision), which may not accurately capture more complex or non-linear patterns in real-world crime data. Decision Trees have also being in CL context, but mainly as a comparison with LR approaches. For example, Tonkin et al.~\cite{tonkin_linking_2019} demonstrated the superior performance of regression models over tree-based models and showcased that incorporating specific crime type behaviors could further enhance performance. The Appendix table (Section \ref{appendix}) provides more examples on how these models have been used.

This study will give more emphasis on more advanced machine learning techniques, particularly because these approaches have not been covered in previous literature surveys. We will also focus on CCA reactive linkages, as this method is preferred among researchers. However, other studies have explored machine learning clustering techniques to evaluate crime linkage decisions \cite{reich2015partially,lin2006outlier,wang2015finding}. Additionally, while not strictly classified as machine learning, some authors have utilized fuzzy logic to assess crime linkage scenarios \cite{stoffel2012fuzzy,goala2019intuitionistic,goala2018fuzzy,albertetti2013crilim}.

\subsection{Machine Learning studies}

With the advancements in Artificial Intelligence, numerous studies have harnessed these technologies for applications within the justice system. It is important to distinguish between crime prediction and crime linkage. Crime prediction, also known as Predictive Policing, has been the focus of numerous authors in Computer Science, employing a wide array of methodologies \cite{mandalapu_crime_2023}. Essentially, crime prediction involves the creation of a model trained on big data, typically to detect crime locations, although it has also been used to predict the likelihood of an individual committing a crime \cite{meijer2019predictive}. In contrast, Crime Linkage is concerned with connecting one crime to another, potentially identifying the same offender. Within the realm of Machine Learning, there also exists a diverse range of methodologies to address the crime linkage problem. 

Machine Learning is the science behind Artificial Intelligence focused on discovering patterns and insights in large and complex datasets. While both statistics and ML share common foundations, they usually differ in their applications \cite{ij2018statistics}. Statistics typically deals with well-defined models and smaller datasets, aiming to make inferences and understand relationships. In contrast, machine learning often handles larger, more intricate datasets, using advanced algorithms to uncover patterns and make predictions. Notably, Natural Language Processing (NLP) techniques, a branch from ML, have been employed on textual data from police reports to discern relationships among incidents and thereby establish case links \cite{zhu_spatiotemporal-textual_2022,solomon_crime_2020,griffard2019bias,li2022thresholds,li2022supervised,li2020novel,li_approach_2019}. Most of the NLP works focused on different techniques to retrieve information and extracting MO features from these police narratives. Additionally, some studies have utilized text vectorization techniques to compute similarity scores. Commonly referred to as embeddings, this approach involves transforming textual information into high-dimensional vectors, thereby capturing latent textual features that can be compared against other reports. 

A summary of data-driven studies on crime linkage can be found in the Appendix (Section \ref{appendix}) . However, below we describe the main works that can be considered purely based on ML.

Chi et al.~\cite{chi_decision_2017} implemented a simple three-layer neural network, with the middle layer comprising two nodes designed to calculate the similarity within input crime features (MOs). The third layer generated the final output similarity score between the two crimes. While the neural network automatically calculated weights to fit the model, the authors proposed the involvement of an experienced human in the loop to adjust the weights as needed. Additionally, they introduced a separability index to prune input features that did not contribute to the linkage output, thus enhancing the process. This approach not only improved the accuracy of linkage analysis but also revealed to analysts which input features were actually related to the crime linkages. The authors noted that this technique was successfully incorporated into a government security office in China.

Y.-S. Li and Qi~\cite{li_approach_2019} introduced a methodology to compare robbery cases in China by employing various steps to measure differences between two crimes. Their approach involved using absolute distance for numerical attributes, Jaccard’s coefficient for categorical attributes,  similarity between Word2Vec \cite{mikolov_efficient_2013} embeddings of selected keywords, and Dynamic Time Warping (DTW) \cite{sakoe1978dynamic} Information Entropy \cite{shannon1948mathematical} for assessing dissimilarities between the narratives of two crimes, also referred to as ``crime processes''.

Zhu and Xie (2021) proposed employing TF-IDF  and Restricted Boltzmann Machine (RBM) \cite{salakhutdinov2007restricted} with regularization on a textual dataset of 911 calls to extract location, time, and MOs \cite{zhu_spatial-temporal-textual_2021}– an approach they initially presented in their earlier work \cite{zhu_crime_2019}. This unsupervised technique enabled them to determine if a crime series had been detected within a pool of crime incidents. However, personal narratives of crimes can inherit bias context \cite{renauer2011examining, cronin_bias-crime_2007} and their method utilizing 911 calls and a bag of words approach may not be the best choice for detecting crime linkage. For instance, the word ``black'' was among the keywords identified by their method to be used as what they considered MO. Moreover, TF-IDF does not take into account the context in which the retrieved keywords were inserted, rendering it susceptible to bias detection towards longer documents \cite{yahav2018comments}. More on textual embedding techniques will be discussed later.


Another comparable study is conducted by Solomon et al.~ \cite{solomon_crime_2020}, in which the authors leverage burglary police reports to extract 40 predefined MO details using fastText~\cite{DBLP:journals/corr/JoulinGBM16} and smooth inverse frequency (SIF)~\cite{arora2017simple} embedding techniques. They then assess the similarity between the two crimes by incorporating the distance and time difference between the two cases into the MO vector. Subsequently, they input these MO vectors, along with the spatial-temporal differences, into a Siamese neural network~\cite{koch2015siamese} to determine the probability of these two inputs being identical. The authors demonstrated the efficacy of their methods in a different language (Hebrew) using narrative text from victims and dialogue between victims and police officers. While their techniques exhibited high performance in burglaries cases where a predefined set of MO may suffice to characterize the crime, it remains unclear whether these methods would also be applicable to crimes against persons~\cite{yaksic2020addressing}.


In a pragmatic approach, Chohlas-Wood and Levine~\cite{chohlas-wood_recommendation_2019} from the New York Police Department (NYPD) developed an application called \textit{Patternizr}. This tool utilizes a blend of structured information, including location, crime subcategory, \textit{modus operandi} details (weapon, victim count, etc.), suspect information (weight, height, etc.), and unstructured data such as crime narrative complaints. The primary objective is not to compute the final crime linkage, akin to the objective of this research, but rather to furnish a list of the most similar cases to glean insights and enhance investigations. However, while the authors assert fairness by excluding the race attribute, other studies have demonstrated that this approach does not eliminate bias \cite{griffard2019bias,martinez-plumed_fairness_2019}. Furthermore, they utilized Word2Vec which has also been demonstrated to carry potential bias in its embeddings \cite{brunet_understanding_2019, manzini_black_2019}.


A significant challenge in crime linkage arises from the imbalance existing in the gold standard \cite{borg_detecting_2014}. Typically, the datasets used to test models exhibit considerable imbalance, with a surplus of non-linked cases compared to linked cases. Given that most methods involve comparing each pair of crimes, the pool of potential linkages is notably small. Consequently, machine learning models tend to favor the majority class, rendering accuracy evaluations problematic, as they predominantly yield correct predictions for the abundant non-linked cases. Using accuracy as the sole evaluation metric can lead to misinterpretation, particularly in such scenarios where there is a significant discrepancy between the two binary classes. While some authors have proposed methods to address this imbalance issue \cite{li_novel_2020, li_supervised_2022}, their efficacy has been limited, with only marginal improvements observed, and primarily addressed the context of robbery cases. We will dive more into this in Section 4.

\section{Crime Linkage Framework} 

\begin{figure}[h]
  \includegraphics[width=\textwidth]{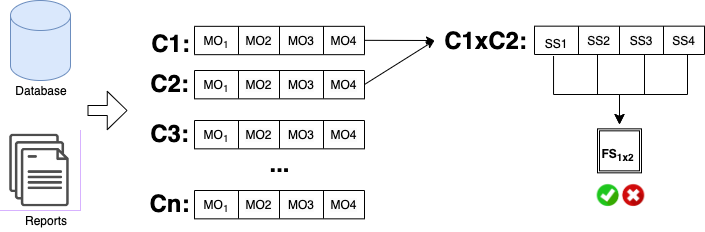}
  \caption
  {
    Typical pipeline of Crime Linkage assessment.
  }
  \label{fig:CL_piline}
\end{figure}

Overall, the crime linkage assessment process follows a similar structure among the studies reviewed. The pairwise comparison is illustrated in Figure~\ref{fig:CL_piline}. The primary goal is to extract variables that characterize each crime case ($C_{1}, C_{2}, ... C_{n}$), which in our context are referred to as MO attributes. Next, the idea is to calculate similarity measurements between these attributes. Finally, all these similarity scores (SS) are averaged (or combined using other methods, such as weighted averaging) to determine a final score (FS). A score above a certain threshold is considered linked, while a score below that threshold is considered unlinked. However, just averaging the scores might not take into consideration other complexities of the criminal information, and thus a machine learning method might be necessary at this stage.

In summary, we can observe three steps in the CL process:

\begin{enumerate}
    \item Identifying MO attributes from each criminal case.
    \item Calculating similarity measures between each pair of crimes.
    \item Using Machine Learning to classify whether this pair is considered linked or unliked.
\end{enumerate}

More details on each step are provided in the following subsections:

\subsection{MO variables}

In this first step, the MO variables are either pre-defined and pulled from a database or extracted from crime narratives or reports. Although techniques vary, most approaches treat the MO as a set of finite attributes that characterize a particular crime. This may explain why most of the work has focused on property crimes, as these types of crimes tend to have more standardized criminal behaviors compared to crimes against persons \cite{fox_what_2018}. This retrieval step is crucial because some level of expertise is necessary to understand which variables influence crime linkage analysis. Human coders, who either insert variables into the database (such as ViCAP \cite{howlett1986violent}) or extract them from police data, are typically subjected to inter-rater reliability assessments to measure agreement in their results. As mentioned in Section 2, this has been a persistent issue in the domain, but it is important to address the significance of reliability in data-driven approaches for crime linkage systems \cite{bennell_computerized_2012}. In Machine Learning method the extraction can be done automatically, typically with the support of Natural Language Processing techniques that retrieves variables from reports, based on textual context or keywords. Lin et al.~\cite{lin2015application} proposed a feature selection process by calculating a index separability in order to prune data to better results in crime linkage. Solomon et al.~\cite{solomon_crime_2020} converted typical police questions into sentence embeddings to find MOs in massive amount of crime narratives.

Common attributes include the location and time of the event, as well as behavioral characteristics such as methods of trespassing, the weapon used, the number of victims, the type of target location, and others. These variables can be numerical, binary, categorical, or based on descriptive words. This step is important because the type of variable dictates the type of similarity measure used in the following step. It is worth noting that a categorical attribute will need to be converted to hot encoding or other sort of numbered format in order to serve as input into a ML algorithm. 

As illustrated in the Appendix table (Section \ref{appendix}), a wide range of MO choices are available for selection in a given model. Nevertheless, location and time, or more precisely, spatiotemporal differences, emerged as some of the most consistent MO characteristics in the literature. Bennell et al.~\cite{bennell2002linking} demonstrated that even using just distance as a factor, without incorporating additional MO features, resulted in high accuracy scores. Similarly, Tonkin et al.~\cite{tonkin_linking_2019} showed that both time and location are sufficient predictors for linking crimes, showing that residential burglaries and commercial robberies tend to happen in regions familiar to the criminal in short lapses of time. Beyond these, there appears to be little consistency among other MO attributes, as they tend to vary significantly depending on the type of crime and the data available.

Many authors thus consider a crime as a vector of MOs. In this way, each crime case can be described as vector C = \{$X_{1}, X_{2}, ... X_{k}$\}, where \textit{X} represents a specific MO or crime feature, and \textit{k} is the total number of features or dimensions. Following this approach, it is essential to map and order every MO attribute correctly. This includes having policies to handle missing values or when certain behaviors are not present. As it will be shown in the next step, each MO needs to be compared pairwise, making the order of the vector crucial, as a similarity cannot be calculated between different types of MOs. For example, it is illogical to compare the location of an event with the time of the event. Some studies have mapped MOs based on the degree of a particular behavior. For instance, a crime analyst might evaluate the level of violence in a particular case. This approach is common when working with fuzzy logic models, where a degree of membership is required to characterize sets of MOs \cite{albertetti2013crilim, goala2018fuzzy, goala2019intuitionistic}.

\subsection{Similarity measures}

As mentioned earlier, similarity scores are used to compare how similar one crime is to another. Most studies use a single method to contrast each feature in the crime vector, while others employ a combination of methods depending on the feature type. The most common measurements are as follows:

\subsubsection{Numerical attributes}

One of the simplest methods for measuring similarity between features, particularly with quantitative attributes, is to use the absolute distance or difference between pairs of attributes. This is rather a dissimilarity score and thus a value close to zero indicates greater similarity. In crime linkage studies, common numerical features are time and location. Researchers often calculate similarity based on the distance between two points in a map (using Euclidean distance, for example) or the difference in time (measured in days, hours, or exact values). This approach is supported by criminology literature, which suggests that crimes committed by the same offender are likely to occur in close proximity both spatially and temporally \cite{tonkin2008link, brantingham2013crime, weisburd2015law}.

\subsubsection{Categorical attributes} 

There are many different was to calculate similarity between categorical values \cite{boriah2008similarity}. The most common technique among CL studies is the Jaccard's coeficient (J) \cite{jaccard1908nouvelles}. Here the coefficient measures not only a single behavior but a collection of them and calculates the ratio between common features and the total number of features (Equation ~\ref{eq:jaccard}). The range of the Jaccard coefficient spans from 0, indicating no common features, to 1, indicating identical vector dimensions. 

\begin{equation}
J(C_1, C_2) = \frac{|C_1 \cap C_2|}{|C_1 \cup C_2|}
\label{eq:jaccard}
\end{equation}
where \( C_1 \) and \( C_2 \) are the sets being compared.

As an example, imagine that we have a case where MOs are ``Entrance by Window'', ``House without Fences'', and ``Criminal Record = Yes'', and another where we have ``Entrance by Door'', ``House without Fences'', and ``Criminal Record = Yes''. Comparing these two cases, Jaccard measurement between them would be 2/3.

Oftenly studies have focused  more on the joint presence of features to characterize behavioral consistency, meaning the same MO variable is found in both crimes. This consideration is important when working with police data, as the absence of a feature does not necessarily indicate it did not occur — it may simply not have been reported due to investigative limitations \cite{tonkin2008link}. However, alternative methods for applying the Jaccard coefficient in crime linkage have been explored. In \cite{tonkin2017using}, the authors tested Bayesian analysis and introduced two additional methods to evaluate performance. One method considered joint absence (Jaccard = 1 if the behavior is either present or absent in both crimes), while the other method treated absence as a new similarity evaluation (Jaccard = 1 if the behavior is present, Jaccard = 2 if absent in both crimes, and 0 otherwise). The authors demonstrated that the latter approach achieved better performance than traditional methods.

Although Jaccard's method is straightforward, it does not account for variations in the levels or degrees within a feature. For example, in burglary scenarios, a ``breaking window'' MO is likely more similar to "entering through the window" than to ``forcing door entry''. However, Jaccard's method does not consider this nuance, treating all features with equal weight regardless of their specific characteristics. Furthermore, police reports may not always provide objective feature selection, which can impact the effectiveness of the method.

Similar to Jaccard, the Dice coefficient (D) (also known as Sørensen-Dice Index) was also found in our pool of studies. The main difference here is that it places more emphasis on the intersection of a pair of crimes. In \cite{ku2014decision}, Dice was used to measure the similarity between two documents by comparing root node attributes. The Dice formula is given by:

\begin{equation}
D(C_1, C_2) = \frac{2 \cdot |C_1 \cap C_2|}{|C_1| + |C_2|}
\end{equation}

Another method for calculating similarity between categorical variables involves constructing a hierarchical tree with domain-specific categories. This approach captures the subtleties of similarity by grouping behaviors that fall under the same domain. For instance, Chi et al.~\cite{chi_decision_2017} developed hierarchical features by adjusting the weights of the leaves based on expert input. By taken into account hierarchical levels of crime behavior, some authors have utilized the taxonomic index (\(\Delta \)s). This approach treats MOs as a tree of parameters, and thus the similarity is calculated considering this tree structure. In fact, Melnyk et al.~\cite{izsak_measuring_2001} compared the two main coefficients for behavioral linkage analysis: Jaccard’s coefficient (J) and the taxonomic similarity index (\(\Delta \)s). They found that while the taxonomic similarity index provides good interpret ability of features, Jaccard’s coefficient, due to its simplicity and sensitivity to the data, is generally more suitable for crime linkage analysis. One disadvantage of hierarchical methods is that it requires some kind of domain knowledge to build the tree and group attributes.

Another interesting method for calculating categorical similarity is based on the frequency of specific features. Goodall's similarity measure (G) operates on the principle that the similarity between two items with the same categorical value should be higher if that value is rare in the dataset. The rationale is that items sharing a rare characteristic are more likely to be similar \cite{wang2013learning}. This technique was utilized in NYPD's Patternizr to analyze structured categorical data \cite{chohlas-wood_recommendation_2019}. According to Wang et al.~\cite{wang2013learning}, a simplified version of Goodall coefficient for attribute \textit{j} can be given by:

\begin{equation}
G_j(C_1, C_2) =
\begin{cases}
1 - \sum_{q \in Q} p_j^2(x) & \text{if } C_{1j} = C_{2j} = x \\
0 & \text{if } C_{1j} \neq C_{2j}
\end{cases}
\end{equation}

where \[
p_j^2(x) = \frac{n_x(n_x - 1)}{N(N - 1)},
\] with \( n_x \) the number of times \( x \) is observed in the collection of \( N \) crimes.

\subsubsection{Cosine Similarity}

Considering NLP studies on crime linkage, a typical similarity measure is the cosine similarity (also found in the literature as normalization inner product or word-to-word similarity), which has been calculated between word or sentence embeddings. Embeddings are the product of the converstion of textual data into multidimensional, using techniques such as Word2Vec \cite{mikolov_efficient_2013}. These vectors capture the semantic relationships between words, and thus, similar words are likely to be close to each other in this multidimensional space. Cosine similarity is then a suitable method to calculate the similarity between two words. This approach can account for word variations existing in categorical attributes, potentially offering a better solution than the Jaccard coefficient in some situations. The Cosine Similarity formula between two vectors, A and B, can be see in Equation ~\ref{eq:cosine_sim}.

\begin{equation}
\text{Cosine Similarity}(A, B) = \frac{A \cdot B}{\|A\| \|B\|} = \frac{\sum_{i=1}^{n} A_i B_i}{\sqrt{\sum_{i=1}^{n} A_i^2} \sqrt{\sum_{i=1}^{n} B_i^2}}
\label{eq:cosine_sim}
\end{equation}

\begin{figure}[h]
\centering
  \includegraphics[width=0.9\textwidth]{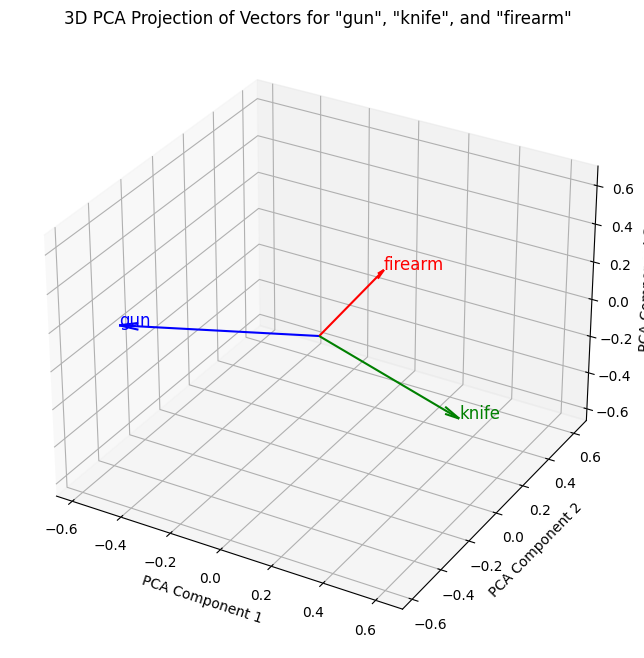}
  \caption
  {
    3 dimensional representation of the words words "gun", "firearm", and "knife". Similarity between 'gun' and 'knife': 0.5895; Similarity between 'gun' and 'firearm': 0.7925; Similarity between 'knife' and 'firearm': 0.5644.
  }
  \label{fig:word2vec}
\end{figure}

As an example, consider the words ``gun'', ``firearm'' and ``knife''. Using the fastText embedding technique \cite{DBLP:journals/corr/JoulinGBM16}, we can plot the three words and calculate the cosine similarity between them (See Figure ~\ref{fig:word2vec}). It is expected that the first two words will be more similar than with the third one, since they share more closely semantic meanings. 

Word embeddings serve as suitable inputs for machine learning models, as they inherently operate on numerical inputs. Various techniques exist for embedding words, including fundamental methods like Bag of Words (BoW) \cite{bagofwords_zhang2010understanding} and Term Frequency-Inverse Document Frequency (TF-IDF) \cite{tfidf_ramos2003using}. However, the popularity of this technique surged following Google's introduction of Word2Vec \cite{mikolov_efficient_2013}. BoW and TF-IDF are well-established techniques in NLP used for analyzing textual documents. BoW represents words in fixed-sized dimensions based on their frequency within a corpus, making it the simplest form of embeddings \cite{bagofwords_zhang2010understanding}. On the other hand, TF-IDF considers words that are frequent within a document but rare across the entire corpus, effectively highlighting important terms while mitigating the impact of common ``stopwords'' such as pronouns, prepositions, and connectors \cite{tfidf_ramos2003using}. However, these methods primarily focus on word frequency and lack the ability to capture contextual relationships within the generated embeddings \cite{qaiser2018text}.

 As mentioned, a significant breakthrough in representing words as numerical vectors came with Word2Vec, developed by Google researchers \cite{word2vec_mikolov2013efficient}. This technique, trained on vast corpora including Wikipedia, Google News Articles, the Web, and Books and Literature, utilizes a shallow neural network. The output of the hidden layer provides word embeddings that capture semantic meaning. Remarkably, this allows for mathematical operations between words, such as ``KING - QUEEN = WOMAN'' \cite{church2017word2vec}. Being a pretrained model, Word2Vec can be used as an index of words and their embeddings. However, despite its revolutionary impact on natural language processing tasks, it is important to note that words embedded using Word2Vec may inherit biases from the texts on which they were trained \cite{manzini_black_2019}.  Manzini et al.~\cite{manzini_black_2019} showed how Word2Vec can make racial associations such as Black-Criminal, Asian-Laborer, and Jew-Greedy.

Our survey analysis revealed that much of the research on CL relies on outdated embedding techniques. While more advanced methods like BERT \cite{devlin2018bert} and large language models (LLMs) \cite{behnamghader2024llm2vec} are now considered state-of-the-art for encoding textual data, we found that many authors still primarily use traditional approaches such as TF-IDF and Word2Vec.

 \subsection{Machine Learning Classification}

\begin{figure}[h]
\centering
  \includegraphics[width=\textwidth]{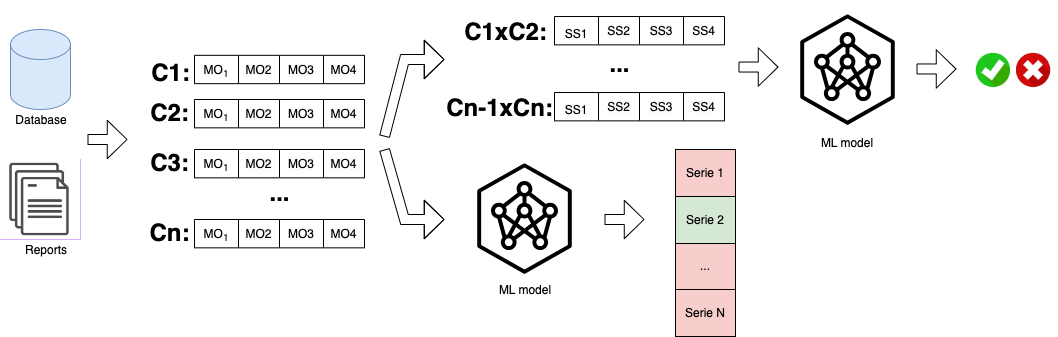}
  \caption
  {
    Typical methodology used in Crime Linkage with Machine Learning algorithms.
  }
  \label{fig:cl_ml}
\end{figure}

The use of machine learning techniques in crime linkage is primarily aimed at classifying whether certain features of a crime indicate a linked crime or associating a crime with a specific crime series, as it can be seen in Figure ~\ref{fig:cl_ml}. These predictions are made after training a chosen algorithm on a large dataset. When data was scarce, simpler algorithms such as linear regression or decision trees were oftenly used. In our study, we observed that most CL approaches input similarity scores between pairs of crimes into the algorithms, rather than using the MO attributes directly. The choice of input features depends on the specific elements the researcher deems relevant for capturing crime patterns. For instance, Chohlas-Wood et al.~\cite{chohlas-wood_recommendation_2019} incorporated historical arrest data alongside similarity scores, allowing the model to learn from this additional information as well.

Below is a brief summary of the most common algorithms found in our literature:

\begin{itemize}
\item {Logistic Regression (LR):} Logistic Regression is a statistical model primarily used for binary classification tasks. It predicts the probability that a given input belongs to a particular class using the logistic function (sigmoid function). This results in outputs that range between 0 and 1, representing class probabilities. It is simple, interpretable, and works well for linearly separable data, making it a popular choice for problems like spam detection and medical diagnosis.

\item {Decision Tree (DT):} A Decision Tree is a non-parametric supervised learning method used for classification and regression. It splits the data into subsets based on feature value tests, creating a tree-like model of decisions and their possible consequences. This model is easy to understand and visualize, as it mimics human decision-making processes. However, decision trees can be prone to overfitting, especially with complex datasets, although they handle both numerical and categorical data effectively.

\item {Random Forest (RF):} Random Forest is an ensemble learning method that constructs multiple decision trees during training and outputs the mode of the classes for classification or the mean prediction for regression of the individual trees. This approach reduces overfitting compared to individual decision trees and enhances model robustness and accuracy. It is particularly effective in handling large datasets and high-dimensional spaces, making it suitable for a variety of applications, from financial modeling to image classification.

\item {Support Vector Machine (SVM)}: Support Vector Machine (SVM) is a supervised learning algorithm used for both classification and regression tasks. It finds the hyperplane that best separates the data into classes in high-dimensional space, with an emphasis on maximizing the margin between different classes. SVMs are effective in high-dimensional spaces and can handle non-linear data through the use of kernel tricks. They are particularly useful in applications like text classification, image recognition, and bioinformatics.

\item {k-Nearest Neighbors (k-NN):} k-Nearest Neighbors (k-NN) is an instance-based learning algorithm used for classification and regression. It classifies a data point based on the majority class of its k-nearest neighbors in the feature space. The simplicity of k-NN makes it easy to implement, but it is sensitive to the choice of k and the distance metric used. It is particularly useful for pattern recognition tasks, such as handwriting digit classification and recommendation systems.

\item {Neural Networks (NN):} Neural Networks are a set of algorithms modeled after the human brain, used for a wide range of tasks, including classification, regression, and complex tasks like image and speech recognition. They consist of layers of interconnected nodes (neurons) that can capture and model complex patterns and non-linear relationships in data. Neural Networks are highly flexible and powerful, but they require large amounts of data and significant computational resources, making them suitable for applications in artificial intelligence and deep learning.

\item {Gradient Boosting Decision Tree (GBDT):} Gradient Boosting Decision Tree (GBDT) is an ensemble learning method that builds multiple decision trees sequentially. Each new tree is trained to correct the errors made by the previous ones, and the trees are combined to make a final prediction. This sequential approach reduces overfitting and increases model accuracy, making GBDT highly effective for a variety of tasks, including ranking, classification, and regression. Careful tuning of parameters like learning rate and the number of trees is essential for optimal performance.
\end{itemize}

In supervised machine learning, the training process requires labeled data. In crime linkage studies, this means that it must be known in advance whether crimes are linked. In practice, this requires that an offender responsible for multiple crimes has already been convicted and that this information is recorded in a relevant database. However, in many cases, a significant portion of the dataset consists of unlabeled data—crimes that are still under investigation. While this unlabeled data can be used to predict whether a crime is linked or belongs to a known offender series, it cannot be used for training. Another major challenge in supervised learning is the class imbalance. Ideally, the model needs examples of both linked and unlinked crimes for training, but in reality, there is often a significant disparity between these two classes, with far more unlinked cases than linked ones. This imbalance can cause the model to favor the majority class, leading to biased predictions. We will explore this issue in greater detail in the following section.

While crime linkage is typically considered as a binary classification problem, Li and Shao \cite{li2022thresholds} demonstrated the suitability of a three-way analysis. In addition to the linked and unlinked classes, they introduced a third category called "boundary". This category encompasses cases that are difficult to classify definitively, which are then referred to a human crime analyst for further evaluation. The challenge lies in minimizing the number of cases that fall into this boundary region, as overloading analysts with too many cases is undesirable, while still maintaining high uncertainty within this region. Increasing the number of cases in the boundary region can enhance the measurement of uncertainty but also places greater demands on human analysis.

Although less common, some studies have approached the crime linkage problem using unsupervised methods. This approach is appealing in the crime linkage context because it addresses the inherent issue of class imbalance and takes out the burden of labeling data. Some studies have explored clustering methods to analyze associations between different crime series \cite{wang2015finding, lin2006outlier, stoffel2012fuzzy, reich2015partially}. However, designing a model to identify connections between crimes without knowing which cases are actually linked presents a big challenge. For instance, Zhu et al. (2022) modeled crime linkage using a Hawkes process, where incidents are governed by an intensity function that decays over time \cite{zhu_spatiotemporal-textual_2022}. This approach, inspired by the behavior of earthquakes, posits that crimes can trigger other crimes, creating a sort of network of linked or associated events. This is somewhat analogous to the predictive policing software PredPol, which employs the Epidemic-Type Aftershock Sequence (ETAS) model \cite{mohler2015randomized}.  

\subsection{Evaluation Metrics}

Within machine learning studies, it is common to separate data to test the trained model. As consequence, the researcher needs to selected an evaluation method to confirm whether his model provides a good performance. The thresholds to determine whether the model had a good performance will depend on the type of crime and the metrics used.  There several evaluation metrics for machine learning and statistics algorithms and this study will present the most commons found in the literature.

The Area Under the Curve (AUC) is the most common evaluation metric in purely statistical CL approaches \cite{ewanation2023receiver}. Also known as the Receiver Operating Characteristic (ROC) curve, it is particularly useful for imbalanced datasets, as it plots the True Positive Rate (TPR) against the False Positive Rate (FPR) \cite{bennell2009addressing,tonkin2017using}. TPR, also known as sensitivity or recall, is the proportion of actual positives that are correctly identified by the model. It is calculated as 
TPR = TP / (TP + FN), where TP represents true positives and FN represents false negatives. FPR, on the other hand, is the proportion of actual negatives that are incorrectly identified as positives by the model. It is calculated as 
FPR = FP / (FP + TN), where FP represents false positives and TN represents true negatives. According to Swets (1988), an AUC over 70\% is considered sufficient for a good performance, however this will most likely depend on the number of data points and type of crime \cite{swets1988measuring}. Although AUC/ROC is the preferable choice among author, many agree that the evaluation process should be tested in more than one technique \cite{tonkin_linking_2019}.

Moving forward to machine learning methods, we have other common evaluation metrics, such as accuracy, precision, recall, and f1-score, specially used when in supervised fashion. Accuracy is one of the most straightforward metrics used to evaluate the performance of a classification model. It is the ratio of correctly predicted instances (both positives and negatives) to the total number of instances. Precision is the ratio of correctly predicted positive observations to the total predicted positives, which thus takes the formula TP / (TP + FP). The F1-score is the harmonic mean of precision and recall, providing a balance between the two, which can be written as 2 x (Precision x Recall) / (Precision + Recall). The choice among these metrics, will depend on how the researcher wants to measure performance. For instance, in cases where the classes are imbalanced, accuracy can be misleading. The F1-score considers both false positives and false negatives, providing a more informative measure of a model's performance.

It is also common to use k-fold cross validation when evaluating a model. In this method, the model is trained and evaluated k times, with each iteration using a different fold as the test set and the remaining k-1 folds as the training set. The performance metrics from each iteration are then averaged to provide a final evaluation measure. This approach aims to balance performance by addressing both bias and variance. A specific case of k-fold cross-validation is Leave-One-Out Cross-Validation (LOOCV), where k is equal to the number of data points in the dataset. While LOOCV can be particularly useful for small datasets, it is computationally expensive due to the high number of iterations required.

\section{Challenges in Crime Linkage}

Applying machine learning to large-scale crime data is valuable for uncovering the nuances and complexities of criminal behavior patterns. However, this field presents several challenges that require careful consideration. We have identified four key challenges associated with working with these approaches. A related work comes from Yang et al.~\cite{yang2021survey}, although their survey is based on a very short pool of papers.

\subsection{Imbalanced Data}

Probably the most common issues encountered in the literature when using big amounts of data to crime linkage is the problem of data imbalance. In crime series analysis, most available data points are typically unlinked. Although recidivism is not rare in criminal behavior, this imbalance can arise for various reasons, such as ongoing investigations where no offender has been identified yet or limitations in police infrastructure, such as jurisdictional constraints, which hinder the ability to connect offenders. Consequently, when preparing the data, there is often a decision to determine some cases as labeled (linked or not) and others as unlabeled. It is anticipated that there will be fewer serial crimes compared to one-off cases. 

In these imbalanced datasets, machine learning models tend to favor the majority class, typically the non-linked cases, due to several factors. First, most algorithms aim to minimize overall error during training, which can lead them to focus on the majority class to achieve higher accuracy. When the vast majority of cases are non-linked, the model can achieve a seemingly high accuracy rate simply by predicting all cases as non-linked, as the model is trained to minimize error across the dataset. This tendency to favor the majority class can result in poor performance when predicting the minority class (linked cases). For example, a model may produce an accuracy of 90\% in a dataset where 95\% of the cases are non-linked, yet it may fail to identify any linked cases at all. This scenario highlights a critical issue: accuracy alone is an inadequate metric for evaluating model performance in imbalanced contexts. Instead, other metrics such as precision, recall, F1-score, and the area under the receiver operating characteristic curve (AUC-ROC) provide a more nuanced understanding of model performance, especially for the minority class.

Based on our review, it appears that earlier studies utilizing logistic regression (LR) models evaluated their methods using the AUC/ROC metric, addressing the imbalance issue by adjusting their false alarm thresholds according to expert opinion \cite{tonkin2017using}. However, these studies typically involved significantly fewer data points compared to more advanced machine learning methods, leaving the problem of data imbalance an open area for further research.

There are simple approaches to addressing the data imbalance problem, such as undersampling the dataset by discarding the surplus of non-linked cases to match the number of linked samples \cite{ali2013classification}. However, this method can result in a dataset so small that the use of machine learning becomes unjustifiable. Other authors have proposed alternative methods to handle the imbalance issue. These include oversampling techniques, such as SMOTE (Synthetic Minority Over-sampling Technique), which generate synthetic samples for the minority class, and cost-sensitive learning, where different misclassification costs are assigned to the majority and minority classes to mitigate the bias toward the majority class \cite{hossain2020crime}. These techniques aim to improve the performance and reliability of machine learning models in the context of imbalanced datasets.


Li et al. (2020) demonstrated that using the Information Granule Random Forest (IGRF) method can reduce the number of non-serial pairs in the dataset \cite{li2020novel}. Their approach involves altering the granularity of the data by clustering similar cases using the k-nearest neighbors algorithm. This allows the model to train on a smaller dataset without significant loss of information from the original data. However, the improvements in performance are modest, with an increase in the true positive rate ranging from 2-5\%.

The same authors later proposed a new solution using combined blocking to address the imbalance problem \cite{li_supervised_2022}. This approach involves creating behavioral keys, which are blocks of cases sharing similar and rare behaviors or MOs. This method results in a new dataset with fewer non-serial crimes. They demonstrated that by retaining 98\% of the serial cases and eliminating 65\% of the pairs, the accuracy, recall, and F1 score improved compared to not using combined blocking. While accuracy did not change significantly, recall and precision showed substantial improvement.

\subsection{Bias}

Although the issue of bias is not extensively discussed in the specific literature, particularly regarding machine learning techniques, we found evidence that this challenge still demands significant attention. Bias within machine learning is a major concern, especially with supervised methods, which can amplify bias during the labeling process. Studies have revealed biases in police data, particularly against minority groups \cite{brantingham2017logic}. Additionally, police underreporting can result in data that does not accurately reflect the true situation \cite{buil2021accuracy}. In the crime linkage process, it is crucial to evaluate whether crime associations are driven by ethnicity or other biased attributes, ensuring that the analysis remains fair and unbiased. Especially, there is the necessity of attention to which type of data is being served as input to ML models.

Bias can be inserted in ML application in any or all of these three stage: preprocessing (trained data), in-processing (withing the algorithm itself), and post-processing (after training) \cite{hort2024bias,friedler2019comparative}. If the police data used to train models for crime likange is biased, then the result will be a biased model. An example of potential preprocessing bias can be found in \cite{zhu_spatial-temporal-textual_2021} where the author used 911 call narratives to train their model, using TF-IDF to extract keywords from the texts.  The word ``black'' was among the keywords identified by their method, which would later to be used as a particular attibute to describe that crime. Although extracting keywords from documents seems a good approach to characterize a document, in a police context, it is important to either debias the dataset first or evaluate how potential biased keywords are affecting your results (and workaround them).

Chohlas-Wood and Levine~\cite{chohlas-wood_recommendation_2019} from the New York Police Department (NYPD) developed an application called \textit{Patternizr}, which they claim is bias-free. This tool utilizes a blend of structured information, including location, crime subcategory, \textit{modus operandi} details (weapon, victim count, etc.), suspect information (weight, height, etc.), and unstructured data such as crime narrative complaints. However, while the authors assert fairness by excluding the race attribute, other studies have demonstrated that this approach does not eliminate bias \cite{griffard2019bias,martinez-plumed_fairness_2019}. Furthermore, they utilized Word2Vec which has also been demonstrated to carry potential bias in its embeddings \cite{brunet_understanding_2019, manzini_black_2019}. Although the tool is not public available, it can be considered an example of both pre and in-processing bias.

The best way to build ML models to support criminal decisions is to make sure they are interpretable and explainable \cite{rudin2019stop}. This way the nuances of how the model operates can be assessed and thus corrected, mitigate or even discarded if necessary. In the era of LLMs and Generative AI it has becoming each day more challenging to explain some of the models behaviors \cite{longo2024explainable}. However, for a practical usuage in the criminal justice it is important to garantee interpretability in order to maintain fair and court acceptable results.

\subsection{Labeled Data}

We mentioned that labeling data can add bias into the model. Besides dealing with this problem, labeling data can also be quite cumbersome. In CL scenarios labeling is required when determining the MO attributes and if cases are connected or not. The stage where MO needs to be extracted to create MO vectors has counted on machine tecnhiques or purely human manual work. The latter case requires some level of expertise to appropriately code the behavior that will feed the model. The code evaluation and reliability in data-driven approaches for crime linkage systems has been topic of extensive research \cite{bennell_computerized_2012}. 

Martineau and Corey~\cite{martineau_investigating_2008} conducted a study to assess the agreement level among Canadian police officers in coding crime linkage scenarios, revealing a 38\% agreement for homicide cases and 25\% for sexual assault cases. This underscores the subjective nature of human evaluation in criminal investigations. Furthermore, the decision-making process for crime linkage itself has sparked debate due to the absence of standardized training or evaluation methods for this complex and subjective task. For example, Bennell et al.~\cite{bennell_linkage_2010} found that students outperformed police personnel in associating crimes with serial offenders, with a logistic regression model outperforming humans. The Belgian version of ViCLAS was also examined in a study by Davies et al.~\cite{davies2018practice}, revealing ongoing improvements in the coding process but highlighting the need for further enhancements, such as standardization of inter-rater evaluations. In another study, Pakkanen et al. (2012) found that student subjects correctly linked 61\% of cases \cite{pakkanen2012effects}, showing a slight difference from the work by Santtila et al.~\cite{santtila2008behavioural}, where the model achieved 63\% accuracy. In a positive note, the former study did not find evidence of bias influencing the coding process.

There is also the problem of each variables to select. Some argue that individual behavior more accurately maps MO for crime linkage. Authors in \cite{salo2013using} advocated for the use of individual behaviors in homicide data, but their study suggested that determining which behaviors are pivotal for crime linkage poses a significant challenge. Interestingly, their findings revealed that nearly identical accuracy levels could be achieved using just 15 out of the 90 behavior categories. Other studies have used grouping tecniques to agregate variables and feed the model with these cluster. Another example with homicides was done by \cite{melnyk_another_2011}. The authors hierarchically categorized 39 "crime scenes behaviors" into five distinct categories: Plan, Control, Ritual, Impulsivity, and Ritual (later subdivided into two groups of organized and disorganized). Interestingly, with this approach Melnyk et al.'s work in \cite{fox_what_2018} achieved a high accuracy rate (96\%) for detecting crime linkage.

\subsection{Crime Type}

The literature on crime linkage encompasses a wide array of methods and evaluation metrics, alongside varied datasets employed to establish connections between criminal incidents. Numerous studies have showcased CL outcomes, particularly in cases involving property crimes and sexual offenses (not necessarily sexual homicides). However, as mentioned, it is notable that the research community has predominantly focused on linking burglaries and robberies, with comparatively less attention directed towards other types of crimes \cite{fox_what_2018}. It is still an open research question of how crime linkage insights can be achieved on other types of crime, especially violent crimes, such as murders.

It is valid to reason that different crime types will have different MO variables. Nevertheless, which variables best maps or characterizes a crime is still under research experiments. Reducing a crime description into a finite and define set of MO attributes is a challenging process, which remains an open area of study. Researchers need to first identify attributes that characterize each individual crime case and then just use a pool of them for the crime linkage goal. As mentioned, most of the CL studies used proprietary crime data, which, due to its nature, can be considered suitable for this mapping behavior into a set of standarized variables.  However, this approach imposes constraints on the breadth of potential criminal behaviors, posing significant challenges, especially in the analysis of violent crimes like homicides \cite{douglas2006modus}. For example, religious ritual killings may display unique MOs that diverge from standardized lists of MO characteristics \cite{hazelwood_relevance_2016}. In way, selecting MOs to a finite and discrete multidimensional vector somewhat contradicts the assumptions adopted fro crime linkage since very rare attributes might not be captured when extracting MOs from criminal data. In \cite{solomon_crime_2020} for example, a set of 40 MOs was particularly searched when applying to cosine similarity to their narratives datasets. Further studies are required to capture the dynamics and flexibility that can shape a criminal behavior.

While this flexibility may add complexity to the CL task, applying machine learning methods could help capture the nuances of how various MO segments influence crime associations. It is important to use models that allow for interpretability, enabling us to understand the impact of different inputs on the outputs. Although domain knowledge remains important, machine learning has the potential to break down the complexity and better understand criminal behavior patterns.

\section{Conclusions}

Our survey highlights how crime linkage studies have evolved with the increasing availability of data. We observed a clear shift from traditional statistical models to more advanced machine learning approaches, driven by the need for more sophisticated techniques to handle the growing data volume. Notably, there has been a recent surge in interest in using crime linkage decisions based on police textual data, such as crime narratives, police reports, and 911 calls. This shift has introduced NLP techniques into the field, attracting the attention of computer scientists to an area previously dominated by criminologists, psychologists, and statisticians.

This study analyzed key works on CL that utilized machine learning methods, exploring how different authors approached the topic. In doing so, we identified a clear framework that was consistently applied across most studies. We broke this framework down into distinct steps, each of which was further detailed to enhance understanding. This structure can help establish a common language among researchers from different fields and encourage broader contributions to the discipline, particularly when applying data-driven methods.

We also observed that the application of machine learning methods to CL is still far from being fully adopted in real-world scenarios. Several key challenges in the literature must be addressed to enable ethical and practical implementation. Four primary challenges stood out: imbalanced data, biased data, the availability and reliability of labeled data, and the crime type influences. While there are certainly more obstacles, these were the most prominent in the context of artificial intelligence approaches. Moving forward, we believe researchers need to be mindful of these challenges and take them into account in their studies to advance the field ethically.

Despite the challenges, our survey of studies shows that machine learning holds great promise for enhancing CL processes. ML offers an efficient way to break down complex information and uncover patterns of criminal behavior that would be difficult or too labor-intensive to identify through human effort alone. As a result, both law enforcement and researchers can benefit from these models by gaining valuable insights for CL decisions and integrating them into police investigations. The diverse approaches observed suggest that collaboration between experts across domains (criminologists, computer scientists, and law enforcement) has the potential to yield more impactful results.

\bibliographystyle{unsrt}  
\bibliography{references}  


\input{appendices.tex}

\end{document}

%% file: appendices.tex

\newgeometry{margin=1cm} 
\begin{landscape}

\section{Appendix} \label{appendix}


\begin{center}

\begin{longtable}{p{1cm}p{2.2cm}p{3.4cm}p{1.3cm}p{3.5cm}p{2cm}p{1.5cm}p{5.5cm}}
\small
\\
\caption{Overview of the main data-driven studies on Crime Linkage.} \label{mylongtablelabel}
\\
\hline
\textbf{Pub. Year} & \textbf{Title} & \textbf{Data used} & \textbf{Type of Crime} & \textbf{MOs used} & \textbf{Similarity Measures} & \textbf{ML technique} & \textbf{Summary} \\ \hline
\endfirsthead

\hline
\multicolumn{3}{@{}l}{\ldots continued}\\\hline
\textbf{Pub. Year} & \textbf{Title} & \textbf{Data used} & \textbf{Type of Crime} & \textbf{MOs used} & \textbf{Similarity Measures} & \textbf{ML technique} & \textbf{Summary} \\ \hline
\endhead 

\hline

2022 & Thresholds learning of three-way decisions in pairwise crime linkage \cite{li2022thresholds} & 364 cases with 111 serial offenses from Zhengzhou City, Henan Province, China. & Robbery & 11 MOs: number of criminals, tools used, way criminals disguise, way victims are harmed, way criminal rob property, way criminal threat victims, part of victim being harmed, item robbed, actions to control victim, way criminals break obstacles, actions taken by criminals & Absolute Distance, Jaccard and Cosine Similarity. & RF & The authors proposed a ternary classification system for linked crimes, rather than the traditional binary approach. In addition to the 'linked' and 'not linked' outcomes, they introduced a middle category where the linkage decision remains uncertain, necessitating further evaluation by criminal experts. Their goal was to optimize the decision boundary, minimizing the number of cases with high uncertainty. \\ \hline

2022 & A supervised machine learning framework with combined blocking for detecting serial crimes \cite{li_supervised_2022} & 364 cases with 111 serial offenses from Zhengzhou City, Henan Province, China. & Robbery & 11 MOs: number of criminals, tools used, way criminals disguise, way victims are harmed, way criminal rob property, way criminal threat victims, part of victim being harmed, item robbed, actions to control victim, way criminals break obstacles, actions taken by criminals. & Absolute Distance, Jaccard and Cosine Similarity. & LR, KNN, GDBT, NN, and RF. &  The authors addressed the class imbalance problem by reducing the number of case pair assessments. Using behavioral key features and combined blocking methods, they first evaluated whether a case pair possessed sufficiently strong characteristics to be considered in the CL decision process. \\ \hline

2022 & Spatiotemporal-textual point processes for crime linkage detection \cite{zhu_spatiotemporal-textual_2022} & 349 burglary and 333 robbery crime incidents. collected from 911 calls from the Atlanta Police Department. 56 incidents were identified as being from serial offenses. & Robbery and Burglary. & 280 TF-IDF keywords (for example: home, door, window, stolen), along with time of the event and police beat location. & Cosine Similarity. & Un-supervised learning, using RBM and EM algorithm. & They reformulate the CL problem as a Hawkes process, where an incident follows an intensity function that decays over time. Similar to earthquakes behavior, the authors considered that crimes has an effect of influence other crimes, which can considered linked or somehow associated. However, the distance between crime is not used as an decay effect, but rather the linkage between crimes are correlated with similar narrative descriptions. \\ \hline

2020 &  A novel random forest approach for imbalance problem in crime linkage \cite{li2020novel} & 364 cases with 111 serial offenses from Zhengzhou City, Henan Province, China. & Robbery & 111 MOs: number of criminals, tools used, way criminals disguise, way victims are harmed, way criminal rob property, way criminal threat victims, part of victim being harmed, item robbed, actions to control victim, way criminals break obstacles, actions taken by criminals. & Absolute Distance, Jaccard and Cosine Similarity. & IGRF & They addressed the imbalance problem by adjusting the data granularity. By applying IGRF, they reduced the number of non-serial pairs through clustering of similar pair-cases. This approach led to improvements of around 2-5\% compared to traditional RF methods. \\ \hline

2020 & Crime linkage based on textual hebrew police reports utilizing behavioral patterns \cite{solomon2020crime} & 65,990 Hebrew reports that occurred in Israel between the years of 2005-2018. 9,622 of the reports are labeled with the criminal identity. & Burglary & Location, time, and 40 MOs (not clearly specified, but the authors "focused on the burglary characteristics (e.g., source of entry) rather than on criminals’ attributes (e.g., visual description, ethnicity)"). & Spatio-temportal difference and GBM to classify how likey a MO is present in a report. & Siamese Neural Network & They presented a solution that extracts MO in a language independent fashion. They used fasText to create the embeddings and the way they extracted MO is similar to how police who interview a victim or suspect. They compared cosine similarities from defined questions and words and sentences generated by the embeddings. The cosine similarities are fed into a GBM that classifies how likely a MO is present in a report. These probabilities and the spatio-temporal difference are added to a SNN that compares two reports and gives a score how close they are from each other. Their solution achieved an f1-score of 92\%. \\ \hline

2019  & An approach for understanding offender modus operandi to detect serial robbery crimes \cite{li2019approach} & 334 cases with 86 serial offenses from Zhengzhou City, Henan Province, China.  & Robbery & 9 MOs: number of criminals, tools used, way criminals disguise, way victims are harmed, way criminal rob property, way criminal threat victims, way criminals break obstacles, actions taken by criminals, crime process. & Absolute distance, Jaccards, Cosine Similarity, and DTW. & LR, SVM, KNN, NN, and RF & In addition to using numeric, categorical, and keyword features, they incorporated crime process data to enhance the model's input. The crime process consists of two sequences: one containing only nouns (objects) and the other only verbs (actions), extracted from narrative reports. The similarity between these sequences is calculated using DTW, and the overall process similarity is determined by the weighted sum of these two similarity measures. The weights are computed using information entropy to optimize the results. They demonstrated that adding process information significantly improved performance, with all classifiers achieving over 90\% accuracy. \\ \hline

2019 & Linking serial sexual offences: Moving towards an ecologically valid test of the principles of crime linkage \cite{woodhams2019linking} & 3,364 sexual offences from 5 different countries (UK, South Africa, Finland, Netherlands, Belgium), where 2,081 were solved serial crimes, 1,191 were solved apparent one-off crimes (n = 1,191), and 91 were unsolved serial crimes that were linked by DNA. & Sexual offenses & 166 MOs related to gain and maintain control over the victim, associated with exiting the crime scene or evading capture, sexual behaviours, target selection variables, and behaviours thought to reflect the offence ‘style’ of the offender and that ‘are not directly necessary for the success of the attack'. & Jaccard. & LR. & They demonstrated that incorporating one-off and unsolved crime data points does not negatively impact crime linkage predictions. However, they acknowledged that the method may produce a notable number of false positives. Despite this, the overall accuracy was strong, particularly with a larger sample size compared to previous studies. \\ \hline

2019 & Linking property crime using offender crime scene behaviour: A comparison of methods \cite{tonkin_linking_2019} & 160 residential burglaries committed by 80 serial offenders in the Greater Helsinki, Finland. 376 vehicle theft crimes committed by 188 serial offender in Northamptonshire, UK. 118 commercial robberies committed by 59 offender in the Greater Helsinki, Finland. & Burglary, Car Theft, and Robbery. & Burglary MOs: location of the crime, time and day of the week, type of property, method of entry, the offender's search behaviour, and the type and cost of property stolen. Car theft data MOs: location, type of car that was stolen, age of the vehicle, time and day of the week, how the vehicle was entered and started, and the physical state in which the vehicle was recovered. Commercial robbery MOs: location, type of business robbed, time of day and day of the week, whether a disguise was worn, weapon use, number of offenders, use of violence, language used, and the type and cost of property stolen. & Jaccard. & LR, DT and Bayes Model. & They demonstrated better performance using Bayesian and regression models, with distance and temporal proximity serving as the primary predictors. They also discussed the limitations of relying solely on AUC for evaluation, suggesting that this could lead to misinformation. As a solution, they proposed using ranked lists of matched crimes to provide more accurate insights. \\ \hline

2019 & A Recommendation Engine to Aid in Identifying Crime Patterns \cite{chohlas-wood_recommendation_2019} & Approximately 30,000 complaints gathered from 10 years of data from the NYPD. & Burglary, Robbery, and Grand Larceny. & 39 MOs: Location features, Time features, Categoriacal features (premise type, crime classification, the M.O. itself, weapon type, and details about the crime’s location), Suspect features, and keywords from complain narratives (TF-IDF). & Mix of similarities measurements (including cosine similarity, Goodall’s similarity, difference in numerical features, count of rare words, count of matches in categorical features). & RF. & This works present the NYPD solution called Patternizr. It basically gives a score of how similar crimes are from each other. The model is trained on a mixture of police complains, structured data (including arrest information). By inserting a complain in the application, it results a list of complains that are similar to the questioned one. \\ \hline

2017 & A decision support system for detecting serial crimes \cite{chi2017decision} & 92 solved cases, committed by 45 offenders (or group of offenders). & Robbery & 22 MOs: Categorical features included number of gang members, somatotype of the suspect, gender of the victim; Numerical features included height of the suspect, age of the suspect, time etc; Hierarchical features included actions taken by the suspects, tools used by the suspects, state of the victim before the crime was committed, etc. & Expert evaluation (categorical attributes), Euclidean distance, and weighted tree distance (hierarchical attributes). & NN. & They proposed a three-layer neural network that outputs a score indicating the likelihood of the input crimes being committed by the same offender. By incorporating a human-in-the-loop approach to adjust the weights for building feature vectors, they achieved a precision of 76\%. Additionally, they used a separability index to prune the input data for better performance. \\ \hline

2017 & Using offender crime scene behavior to link stranger sexual assaults: A comparison of three statistical approaches \cite{tonkin2017using} & 3,364 sexual offences from 5 different countries (UK, South Africa, Finland, Netherlands, Belgium), where 2,081 were solved serial crimes, 1,191 were solved apparent one-off crimes, and 91 were unsolved serial crimes that were linked by DNA. & Sexual offenses & 166 MOs related to gain and maintain control over the victim, associated with exiting the crime scene or evading capture, sexual behaviours, target selection variables, and behaviours thought to reflect the offence ‘style’ of the offender and that ‘are not directly necessary for the success of the attack'. & Jaccard. & LR, DT, and Bayes Model. & Their goal was to compare the three methods, with the Iterative Classification Tree showing slightly better performance, although all methods produced good AUC scores. They also evaluated performance by setting a false alarm cut-off threshold at 15\%, which led to a decrease in AUC. Additionally, they demonstrated that the choice of Jaccard’s design method significantly influences performance. \\ \hline

2016 & A statistical approach to crime linkage \cite{porter_statistical_2016} & 4681 solved breaking and entering crimes reports provided by the Baltimore County Police Department, where 772 were considered serial. & Burglary & 9 MOs: distance, temporal proximity, property type, point of entry, method of entry. & Numerical differences and matching attributes. & LR, Naive Bayes, Boosted tree. & This study demonstrated how to apply the Bayes Factor (BF) to evaluate crime linkages. The key idea is that BF calculates the ratio of probabilities, favoring the linked hypothesis over the unlinked one. They also employed BF along with hierarchical clustering to identify crime series. Additionally, BF was used to rank suspects within crime series, showcasing its versatility in crime analysis. \\ \hline

2015 & Testing the assumptions of crime linkage with stranger sex offenses: A more ecologically-valid study \cite{slater2015testing} & 50 serial and 50 one-off cases from UK. & Sexual offenses & 217 MOs, which included crime scene location descriptions, how the offender approached the victim, verbal themes, and sexual acts performed. & Jaccard & LR. & TThe primary contribution of this work at the time was the inclusion of one-off cases. Surprisingly, this enhancement led to improved model performance compared to previous studies. \\ \hline

2014 & A decision support system: Automated crime report analysis and classification for e-government \cite{ku2014decision} & 100 crime reports that were synthetic generated by non-experts after watching videos. Same crime case could have been written by different people. & Burglary, Robbery, Theft, and Assault. & MOs are derived from the reports using an ontology created in previous work, containing 20 semantic trees, which including 38,000+ keywords and phrases. & Jaccard and Dice. & LR and Naive Bayes. & Similar to CL, their goal was to determine whether one report matches another. Their motivation stemmed from anonymous crime reports, which provide the police with an additional source of crime data but often contain duplicates. They demonstrated that the Naive Bayes algorithm achieved an accuracy of around 94\% in identifying duplicate reports. \\ \hline

2013 & Using Bayes’ theorem in behavioural crime linking of serial homicide \cite{salo2013using} & 116 serial cases collected from Italian newspaper, Internet, and microfilms of library journals. This dataset had a total of 19 series. & Homicide & 92 MOs, not specifically defined, but containing "offence-related information, victim characteristics as well as situational variables". & N/A & Bayes Model. & They achieved a high accuracy of 84\% in linking a crime case to a series. Additionally, they demonstrated that reducing the number of features to 15 still maintained strong performance. \\ \hline

2013 & Comparing the predictive accuracy of case linkage methods in serious sexual assaults \cite{winter2013comparing} & 219 sexual offenses from 1976 to 2009 in UK, in which 90 are serial cases (30 offenders) and 129 are one-off cases. & Sexual offenses & 46 crime scene characterists were coded by experts. Also, they compared their results using Mokken scale to create 7 higher dimensions. & N/A & Bayes Model. & The aim of this study was to compare the accuracy of serial offender linkage when one-off cases are included. The results demonstrated a reduction in accuracy when these cases are added. Additionally, the study found that the Naive Bayes classifier outperformed discriminant function analysis (DFA), using the Mokken scale, achieving an accuracy rate of 34.5\%. \\ \hline

2012 & A Comparison of Logistic Regression and Classification Tree Analysis for Behavioural Case Linkage \cite{tonkin2012comparison} & 376 serial car thefts committed in the UK and 160 serial residential burglaries committed in Finland. & Burglary and Car theft. & 166 Burglary MOs: target characteristics , entry behaviours, internal behaviours, property stolen, the intercrime distance, and combined behavioural domains. 89 Car theft data MOs: target selection choices, target acquisition behaviour,  disposal behaviour , the intercrime distance, and a combined behavioural domain. & Jaccard and intercrime distance. & LR and DT. & Their main findings shoe that LR and DT demonstrated similar accuracy results, but there is evidence suggesting that the Decision Tree model may have overfitted. \\ \hline

2012 & A test of case linkage principles with solved and unsolved serial rapes \cite{woodhams2012test} & 119 cases of sexual assault gathered by the South African Police, with 22 serial offenders and 14 unsolved cases. & Sexual offenses. & Various MO characteristis coded by the author, based on a checklist of traditional rapist behavior (not clearly detailed). & Jaccard & LR. & Previous concerns about generalizing findings from convicted serial offenders to unsolved cases were addressed in this study. Crime pairs identified by MO were more behaviorally similar than those identified through DNA matches, though the effect was small. \\ \hline

2012 & The linking of burglary crimes using offender behaviour: Testing research cross‐nationally and exploring methodology \cite{tonkin2012linking} & 538 solved serial and non-serial residential burglary crimes committed in the Greater Helsinki region of Finland. & Burglary. & 87 MOs in the domains of target characteristics , entrybehaviours, internal behaviours, property stolen, location, time, and a combined behavioural domain. & Jaccard, intercrime distance, temporal proximity. & LR. & The authors demonstrated that the features used for CL in the UK do not necessarily apply to Finland. They showed that incorporating a broader range of behaviors led to better performance in linking crimes within the Finnish context. \\ \hline

2012 & Linking personal robbery offences using offender behaviour \cite{burrell2012linking} & 166 solved offences committed by 83 offenders in UK (Northamptonshire Police). & Robbery. & 48 MOs in the domains of target selection, control, property type, location, and time. & Jaccard, intercrime distance, temporal proximity. & LR. & The author sought to test CL assumptions using a personal robbery dataset, where offenders operated within a confined geographical area. \\ \hline

2012 & The course of case linkage never did run smooth: A new investigation to tackle the behavioural changes in serial car theft \cite{davies2012course} & 258 pairs, which 129 were linked pairs and 129 randomly generated unlinked pairs from UK (Northamptonshire Police). & Car Theft. & 40 MOs in the domains of type of car, age of car, time of day, day of week, type of car theft, innocent party selection, immobiliser present, method of entry, method of starting car, property taken, and vehicle recovery state. & Jaccard. & LR. & The authors aimed to apply CL to a theft dataset, using behavioral features that differed from those in previous research. They also demonstrated that the way certain attributes are manipulated by researchers can significantly impact the effectiveness of the linking process. \\ \hline

2011 & Linking different types of crime using geographical and temporal proximity \cite{tonkin2011linking} & 1,951 crimes committed by 537 offenders from UK. & Various crime types. & Crime time and location information. & Intercrime distance and temporal proximity. & LR. & The authors sought to evaluate CL performance across various crime types and categories, recognizing that the same offender may commit different kinds of crimes. They demonstrated that it is possible to distinguish between linked and unlinked cases, even when dealing with a wide range of crime types. \\ \hline

2010 & Linkage analysis in cases of serial burglary: Comparing the performance of university students, police professionals, and a logistic regression model \cite{bennell_linkage_2010} & 38 pairs of commercial burglaries that occured in a city in UK, of which 8 were considered linked. & Burglary. & 5 MOs: offense locations, the distance between two offenses, entry methods, target characteristics, and property stolen. & Intercrime distance. & LR. & The goal of this paper is not to focus on CL itself, but rather to explore the complexity of the decision-making process involved in CL. They demonstrated that LR achieved better performance than CL decisions made by police officers and trained students. \\ \hline

2010 & Linking serial residential burglary: Comparing the utility of modus operandi behaviours, geographical proximity, and temporal proximity \cite{markson2010linking} & 160 solved cases from UK, 80 linked cases and 80 unlinked cases (Northamptonshire Police). & Burglary. & 79 MOs in the domains of target selection, entry behaviours, and property, as credit card, gifts, mobile phone, and personal music player information. & Jaccard, intercrime distance, temporal proximity. & LR. & This study aimed to compare which input information was most effective for the CL process: MO behavioral attributes, geographic proximity, or temporal proximity. The results showed that geographic distance outperformed the other features in terms of linking accuracy. \\ \hline

2008 & To link or not to link: a test of the case linkage principles using serial car theft data \cite{tonkin2008link} & 386 crimes from the Northamptonshire police force in the East Midlands, UK, which had 193 offenders. & Car theft. & 27 MOs: type of car, age, time of the day, day of the week, method of entry, method of starting car, property taken, vehicle recovery state, and also intercrime distance and interdump distance. & Jaccard and intercrime distance. & LR. & The main objective of their work was to evaluate crime linkage in theft data, which they demonstrated to perform well. They also identified that intercrime and interdump distances were the features most influential in the predictions. \\ \hline

2007 & An empirical test of the assumptions of case linkage and offender profiling with serial commercial robberies. \cite{woodhams2007empirical} & 160 commercial robberies committed by 80 offenders in UK. & Robbery & 71 MOs in the domains of target selection, planning, control, proprety. & Jaccard and intercrime distance. & LR. & The study examined CL in commercial robberies, aiming to test three hypotheses: behavioral consistency, offender behavioral distinctiveness, and homology (a direct relationship between behavior and offender characteristics). The results indicated that further research is needed to support the homology hypothesis. \\ \hline

2005 & Between a ROC and a hard place: A method for linking serial burglaries by modus operandi. \cite{bennell_between_2005} & 634 commerial burglaries and 660 residential burglares collected from UK, with overall 108 serial offenders. & Burglary & MOs in the domains of entry behavior, target characteristics, and items stolen. & Jaccard and intercrime distance. & LR. & The study's primary finding was that inter-crime distance proved more effective than traditional MO indicators for linking burglaries. \\ \hline

2002 & Linking commercial burglaries by modus operandi: tests using regression and ROC analysis \cite{bennell2002linking} & 86 solved cases, with 43 serial offenses from metropolitan UK police force. & Burglary & 4 MOs: intercrime distance entry behaviours, target selection choices and property stolen. & Jaccard and intercrime distance. & LR. & One of the earliest studies on CL. They demonstrated that intercrime distance is a more effective predictor of linkage than other variables. \\ \hline
 
\end{longtable}
\end{center}

\end{landscape}
\restoregeometry